\documentclass[sigconf, screen]{acmart}

\acmYear{}
\acmISBN{}
\acmDOI{}
\renewcommand\footnotetextcopyrightpermission[1]{}
\setcopyright{none}
\usepackage{balance} 
\usepackage{color}
\usepackage{xcolor}
\usepackage{graphicx}
\usepackage{subcaption}
\usepackage{array}

\definecolor{ForestGreen}{RGB}{34,139,34}
\definecolor{RoyalBlue}{RGB}{65,105,225}
\definecolor{Maroon}{RGB}{128,0,0}
\definecolor{NavyBlue}{RGB}{0,0,128}
\definecolor{RedOrange}{RGB}{255,69,0}
\definecolor{OrangeRed}{RGB}{255,69,0}
\definecolor{Purple}{RGB}{128,0,128}
\definecolor{Violet}{RGB}{238,130,238}
\definecolor{Brown}{RGB}{165,42,42}
\definecolor{Pink}{RGB}{255,192,203}
\definecolor{Cyan}{RGB}{0,255,255}
\definecolor{Magenta}{RGB}{255,0,255}
\definecolor{Yellow}{RGB}{255,255,0}
\definecolor{Lime}{RGB}{0,255,0}
\definecolor{Olive}{RGB}{128,128,0}
\definecolor{Teal}{RGB}{0,128,128}
\definecolor{Silver}{RGB}{192,192,192}

\definecolor{gray1}{RGB}{210,210,210}
\definecolor{gray2}{RGB}{50,50,50}
\definecolor{green}{HTML}{2ECC71}
\definecolor{blue}{HTML}{3498DB}
\definecolor{red}{HTML}{E74C3C}
\definecolor{orange}{HTML}{F39C12}
\definecolor{gray}{HTML}{95A5A6}   
\definecolor{Instruments}{HTML}{BB4806}
\definecolor{Rhexis}{HTML}{DBe734}
\definecolor{Pupil}{HTML}{E63377}
\definecolor{Iris}{HTML}{A04feb}
\usepackage{tikz}
\usepackage{multirow}
\usepackage{pifont}
\newcommand{\cmark}{\ding{51}}  
\newcommand{\xmark}{\ding{55}}  
\newcommand{\DP}[2]{%
  \begin{tikzpicture}
    \fill[color=#2]   (0.0 , 0.0) rectangle (#1*6.2ex , 2ex );
  \end{tikzpicture}%
}
\AtBeginDocument{%
  }

\newcommand{\DrawPercentageBar}[1]{%
  \begin{tikzpicture}
    \fill[color=gray2]   (0.0 , 0.0) rectangle (#1*0.04ex , 1.5ex );
    \fill[color=gray1] (#1*0.04ex  , 0.0) rectangle (4.0ex, 1.5ex);
  \end{tikzpicture}%
}

\newcommand{\NP}[1]{
  #1 \DrawPercentageBar{#1}
}
\usepackage{calc}

\usepackage{stackengine}
\newcommand\clapp[3][0pt]{\stackengine{0pt}{#3}{\kern#1#2}{O}{c}{F}{F}{L}}

\begin{document}

\title{WetCat: Enabling Automated Skill Assessment in Wet-Lab Cataract Surgery Videos}

\author{Negin Ghamsarian}
\email{negin.ghamsarian@unibe.ch}
\orcid{0000-0002-0908-8972}
\affiliation{%
  \institution{University of Bern}
  \city{Bern}
  \country{Switzerland}
}

\author{Raphael Sznitman}
\email{raphael.sznitman@unibe.ch}
\orcid{https://orcid.org/0000-0001-6791-4753}
\affiliation{%
  \institution{University of Bern}
  \city{Bern}
  \country{Switzerland}
}

\author{Klaus Schoeffmann}
\email{Klaus.Schoeffmann@aau.at}
\orcid{https://orcid.org/0000-0002-9218-1704}
\affiliation{%
  \institution{University of Klagenfurt}
  \city{Klagenfurt}
  \country{Austria}
}

\author{Jens Kowal}
\email{Jens.Kowal@unibe.ch}
\orcid{https://orcid.org/0009-0007-4936-6141}
\affiliation{%
  \institution{University of Bern}
  \city{Bern}
  \country{Switzerland}
}

\date{}

\renewcommand{\shortauthors}{Negin Ghamsarian, Raphael Sznitman, Klaus Schoeffmann, and Jens Kowal}

\begin{abstract}
To meet the growing demand for systematic surgical training, wet-lab environments have become indispensable platforms for hands-on practice in ophthalmology. Yet, traditional wet-lab training depends heavily on manual performance evaluations, which are labor-intensive, time-consuming, and often subject to variability. Recent advances in computer vision offer promising avenues for automated skill assessment, enhancing both the efficiency and objectivity of surgical education. Despite notable progress in ophthalmic surgical datasets, existing resources predominantly focus on real surgeries or isolated tasks, falling short of supporting comprehensive skill evaluation in controlled wet-lab settings. To address these limitations, we introduce WetCat, the first dataset of wet-lab cataract surgery videos specifically curated for automated skill assessment. WetCat comprises high-resolution recordings of surgeries performed by trainees on artificial eyes, featuring comprehensive phase annotations and semantic segmentations of key anatomical structures. These annotations are meticulously designed to facilitate skill assessment during the critical capsulorhexis and phacoemulsification phases, adhering to standardized surgical skill assessment frameworks. By focusing on these essential phases, WetCat enables the development of interpretable, AI-driven evaluation tools aligned with established clinical metrics. This dataset lays a strong foundation for advancing objective, scalable surgical education and sets a new benchmark for automated workflow analysis and skill assessment in ophthalmology training.
The dataset and annotations are publicly available in Synapse (\url{https://www.synapse.org/Synapse:syn66401174/files/}). 
\end{abstract}


\keywords{Surgical Skill Assessment, Cataract Surgery, Surgical Phase Recognition, Semantic Segmentation, Surgical Workflow Analysis, Computer-Assisted Interventions, Wet-Lab Cataract Surgery, Cataract Surgery Dataset.}

\maketitle

\begin{figure}[tbp]
    \centering
    \resizebox{1\columnwidth}{!}{%
    \centering
        \includegraphics[width=1\columnwidth]{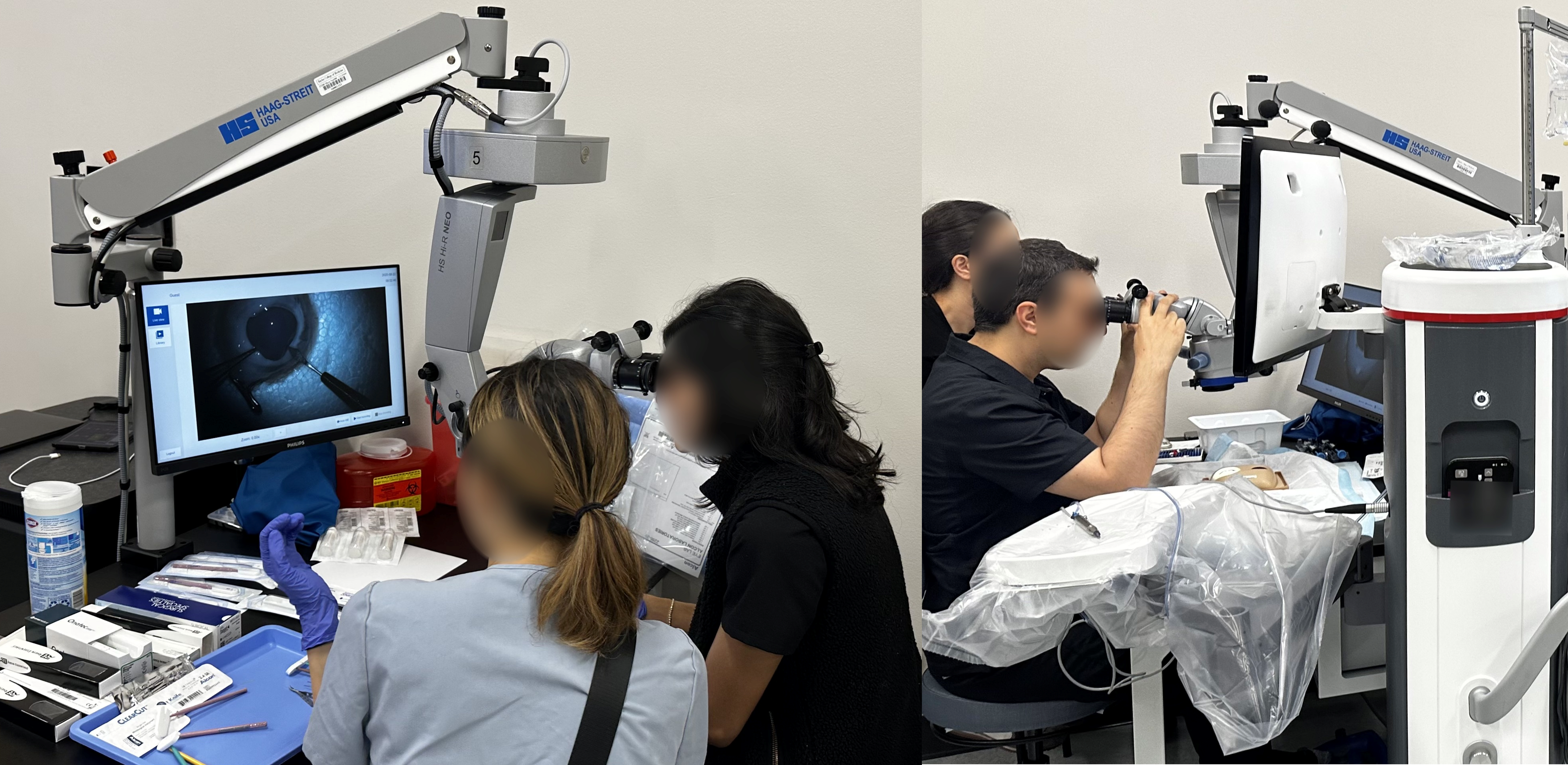}
    }
    \caption{Hands-on cataract surgery training in the wet lab.}
    \Description{Hands-on cataract surgery training in the wet lab.}
    \label{fig:Wet-Lab_environment}
\end{figure}
\section{Introduction}

Cataract surgery is one of the most commonly performed surgical procedures worldwide, with approximately four million cases annually in the United States and around 20 million globally \cite{mcdermott2022overview,rossi2021cataract}. Driven by an aging global population, the demand for cataract surgery is expected to rise significantly, with an anticipated increase of up to 128\% by 2036. Concurrently, the World Health Organization projects that cataract-related blindness will affect nearly 40 million individuals by 2025 \cite{VB2020}. As a result, mastery of cataract surgery has become a critical component of ophthalmology training programs, which increasingly emphasize the need for systematic evaluation and improvement of surgical skills among trainees.

Advances in surgical education technologies have led to the integration of simulation-based training within ophthalmology curricula. Wet-lab environments, in particular, have emerged as essential platforms for hands-on surgical practice, offering anatomically realistic models without the risks associated with patient care (Figure \ref{fig:Wet-Lab_environment}). Unlike dry-lab simulations, which often incorporate embedded sensors to facilitate objective performance evaluation, wet-lab training primarily relies on manual assessment by expert reviewers, a process that is labor-intensive, costly, and frequently lacking in precision and detailed feedback. In this context, automated skill assessment through computer vision presents a promising solution, offering scalable and objective evaluation methods that can enhance both training efficiency and educational outcomes.

Despite the critical role of wet-lab training in developing surgical competencies, current publicly available datasets are derived from real patient surgeries \cite{ghamsarian2024cataract} 
and in most cases focus on isolated tasks such as instrument detection~\cite{al2019cataracts}, phase recognition~\cite{LocalPhase}, or segmentation of anatomical structures and instruments~\cite{CaDIS}. Other datasets target specific objectives, including irregularity detection~\cite{LensID, ghamsarian2023predicting} and relevance-based video compression~\cite{ghamsarian2020relevance}. Accordingly, a considerable body of research has focused on content analysis in real-world cataract surgery \cite{nasirihaghighi2025dual,ghamsarian2024deeppyramid+, ghamsarian2022deeppyramid,marafioti2021catanet,hu2024ophclip,ghamsarian2025feedback,nasirihaghighi2025data,9098318,RBE,hu2024ophnet}.  While these resources have advanced computer vision applications in ophthalmic surgery, they are not designed for comprehensive, automated skill assessment, particularly under the controlled conditions of wet-lab environments. Furthermore, datasets based on real surgeries present significant domain adaptation challenges when applied to wet-lab settings \cite{ghamsarian2023domain}.

Recognizing these gaps, we introduce \textbf{WetCat}, the first dataset of wet-lab cataract surgery videos specifically curated for automated skill assessment. WetCat consists of high-resolution recordings of cataract procedures performed on artificial eyes by trainee ophthalmologists. Each video is meticulously annotated with surgical phase labels and semantic segmentations of key anatomical structures, following standardized assessment frameworks such as GRASIS~\cite{cremers2005global} and OSCAR~\cite{golnik2013development}. The dataset focuses on the critical phases of capsulorhexis and phacoemulsification, which are essential to cataract surgery training. The standardized, reproducible nature of wet-lab procedures further enhances WetCat’s utility for benchmarking and developing AI-driven evaluation tools.
In addition to supporting conventional video analysis tasks such as phase recognition and instrument tracking, WetCat enables objective and interpretable assessments of surgical proficiency. By aligning with established clinical skill metrics, it paves the way for intelligent, data-rich training systems aimed at improving the quality, consistency, and scalability of ophthalmic surgical education.

The remainder of this paper is organized as follows. Section~\ref{sec:methods} reviews the skill assessment criteria that motivate the need for our phase annotations and surgical scene segmentations. Section~\ref{sec: dataset} describes the WetCat dataset in detail. Section~\ref{sec:validation} presents benchmarking and experimental validations of the annotations. Finally, Section~\ref{sec: conclusion} summarizes the work and concludes the paper.

\begin{figure}[tbp]
    \centering
    \resizebox{1\columnwidth}{!}{%
    \centering
        \includegraphics[width=0.8\columnwidth]{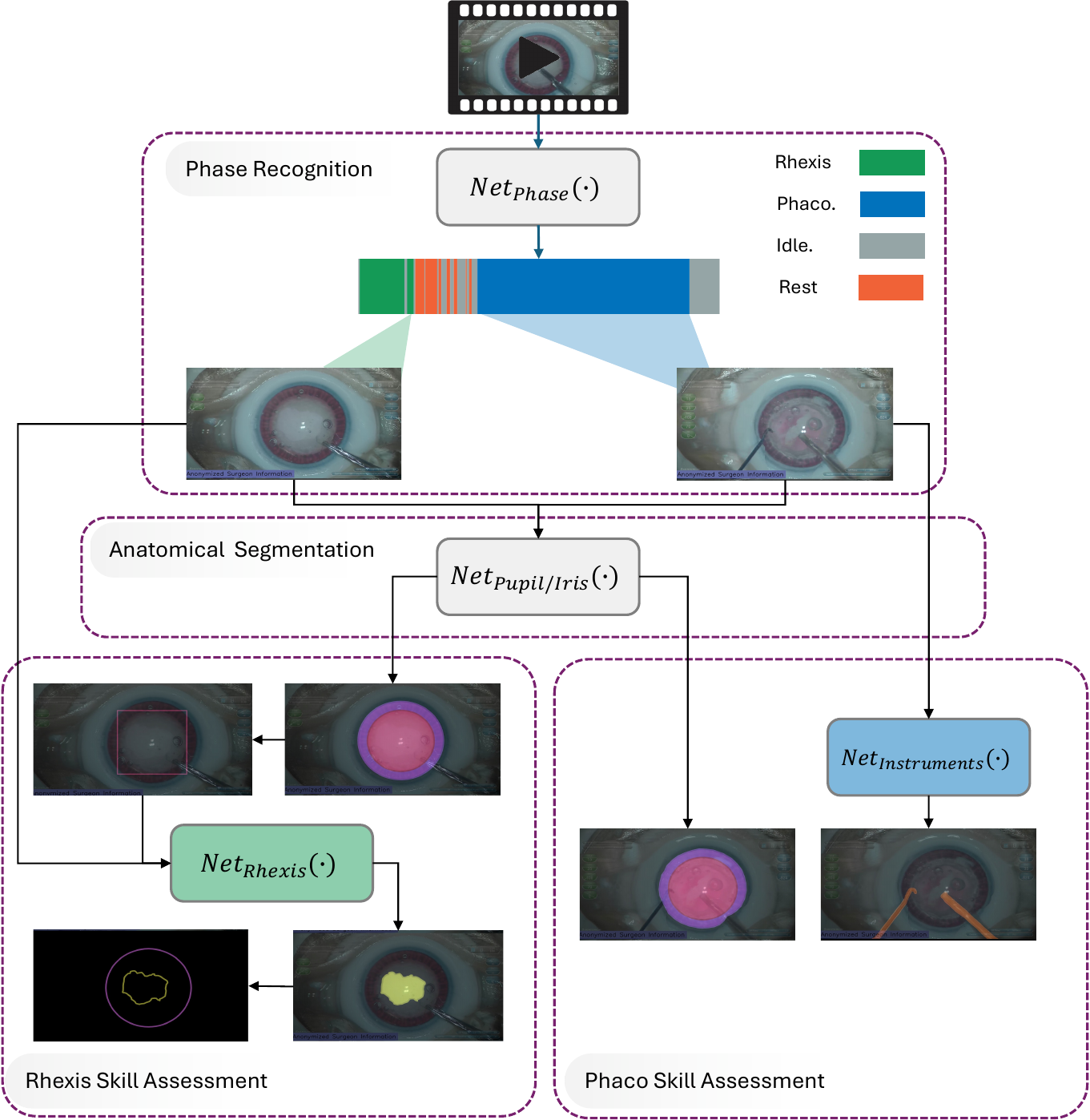}
    }
    \caption{Overall framework for skill assessment in wetlab cataract surgery.}
    \Description{Sample frames from relevant phases in a wet-lab cataract surgery.}
    \label{fig:block-diagram}
\end{figure}

\section{Methods}
\label{sec:methods}

In this section, we present the evaluation metrics for two critical phases of cataract surgery: capsulorhexis and phacoemulsification. These skill metrics are derived from established standards, including GRASIS~\cite{cremers2005global} and OSCAR~\cite{golnik2013development}, ensuring clinical relevance and standardization. The dataset annotations are structured in alignment with these metrics to facilitate objective skill assessment. Figure~\ref{fig:block-diagram} illustrates the overall framework for skill assessment, highlighting the role of phase recognition and semantic segmentation in evaluating surgical skill during the two targeted phases.

\subsection{Capsulorhexis Assessment Metrics}
Capsulorhexis is a precision-critical step in cataract surgery, where the quality of the circular opening in the anterior capsule directly affects surgical outcomes. Its evaluation relies on a combination of shape- and position-based metrics.

\begin{itemize}
    \item \textbf{Roundness}: Measured by the circularity ratio, which compares the object’s area to the square of its perimeter, with values near 1 indicating a nearly perfect circle.

    \item \textbf{Centration}: Assesses alignment accuracy using a centration metric, defined as the Euclidean distance between the capsulorhexis and limbus centers normalized by the limbus radius.

    \item \textbf{Diameter}: Determined as the maximum extent of the capsulorhexis and compared against the clinically ideal range of 4.5--5.5~mm.
    
    \item \textbf{Smoothness}: Evaluates the continuity and regularity of the capsulorhexis edge using curvature-based or Fourier descriptors to quantify local irregularities.
\end{itemize}

To benchmark surgical precision, the detected capsulorhexis can be compared with a \textbf{reference region} defined by ideal geometric criteria, a circular shape with a diameter between 4.5 and 5.5~mm and perfect centration relative to the limbus. Similarity is quantified using the Dice Score and Jaccard Index. Visual overlays further illustrate discrepancies in shape, size, and position.

\subsection{Phacoemulsification Assessment Metrics}
For the phacoemulsification phase, the skill level can be evaluated using ocular stability and procedural efficiency:
\begin{itemize}
    \item \textbf{Eye Stability}: Assessed by tracking the limbus center throughout the phase. Deviations from the initial position are expressed as a percentage of the limbus diameter, with deviations within \(\pm 10\%\) considered acceptable. Larger displacements may indicate inadequate fixation or excessive instrument force.
    
    \item \textbf{Phacoemulsification Duration}: Measures the total time spent in this phase, recorded in seconds, and compared against established clinical benchmarks.
    
    \item \textbf{Non-Dominant Hand Instrument Analysis}: Focuses on tracking the thinner instrument typically used by the non-dominant hand. This analysis evaluates the difficulty of incision handling by assessing the frequency and amplitude of movements, as well as detecting irregular trajectories that may indicate challenges in maneuvering.
\end{itemize}

Together, these metrics provide a comprehensive assessment of performance during two of the most technically demanding phases of cataract surgery.

\begin{figure}[t!]
    \centering
    
    \begin{subfigure}[b]{\columnwidth}
        \centering
        \includegraphics[width=\linewidth]{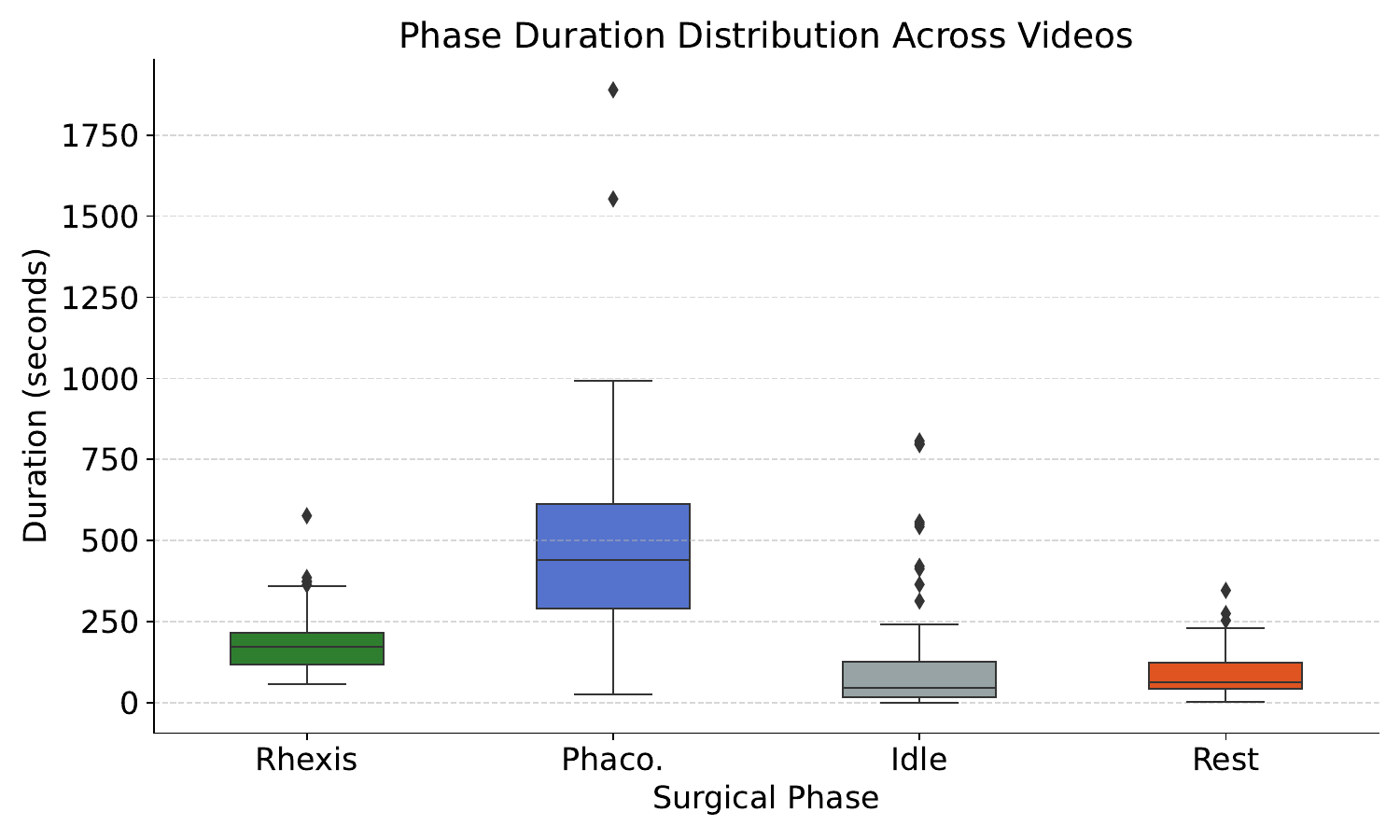}
    \end{subfigure}
    
    \vspace{1em}  

    \begin{subfigure}[b]{0.8\columnwidth}
        \centering
        \includegraphics[width=\linewidth]{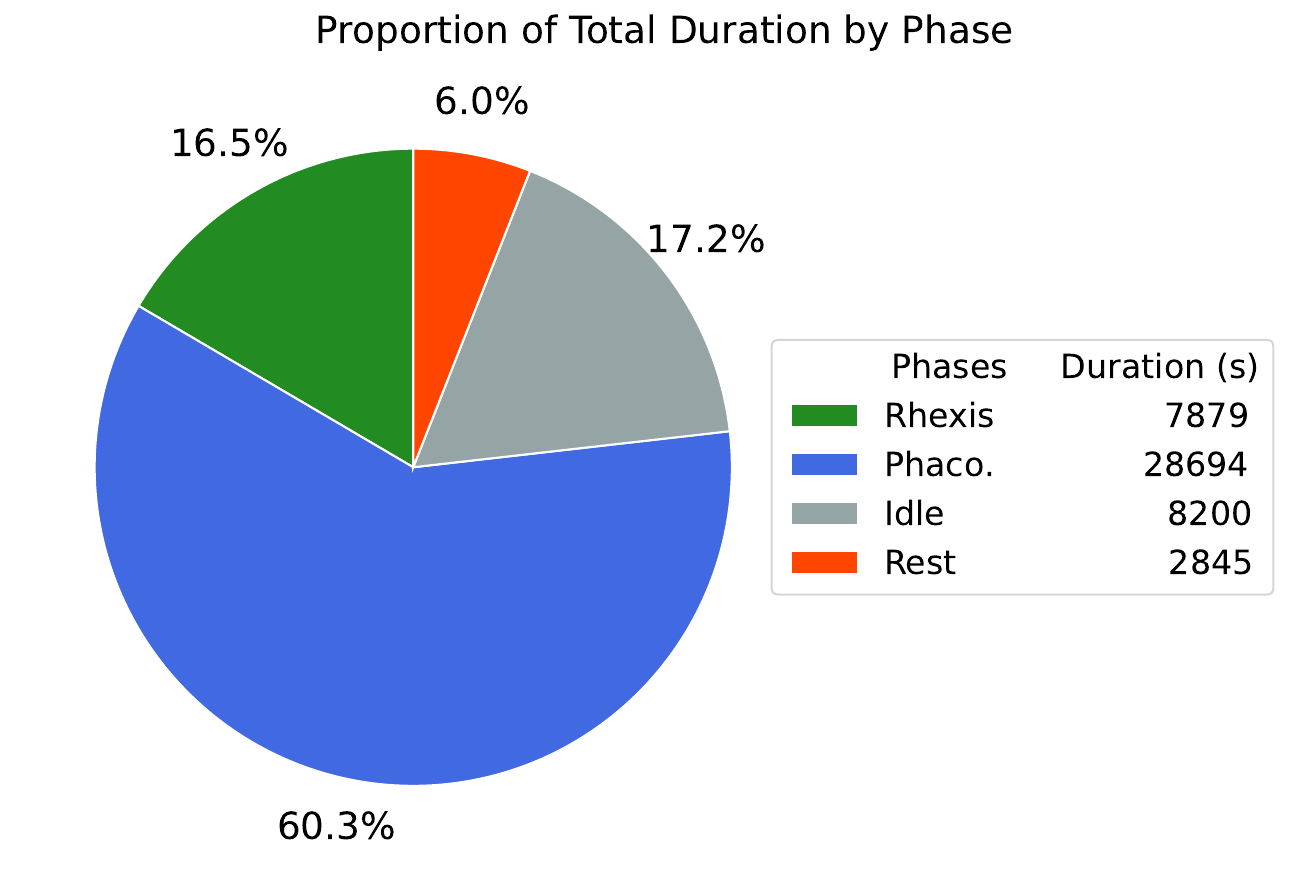}
    \end{subfigure}

    \caption{Distribution of surgical phase durations across videos and overall phase proportions in the dataset.}
    \Description{Distribution of surgical phase durations across videos and overall phase proportions in the dataset.}
    \label{fig:phase_statistics}
\end{figure}

\begin{figure}[t!]
    \centering
    
    \begin{subfigure}[b]{0.97\columnwidth}
        \centering
        \includegraphics[width=\linewidth]{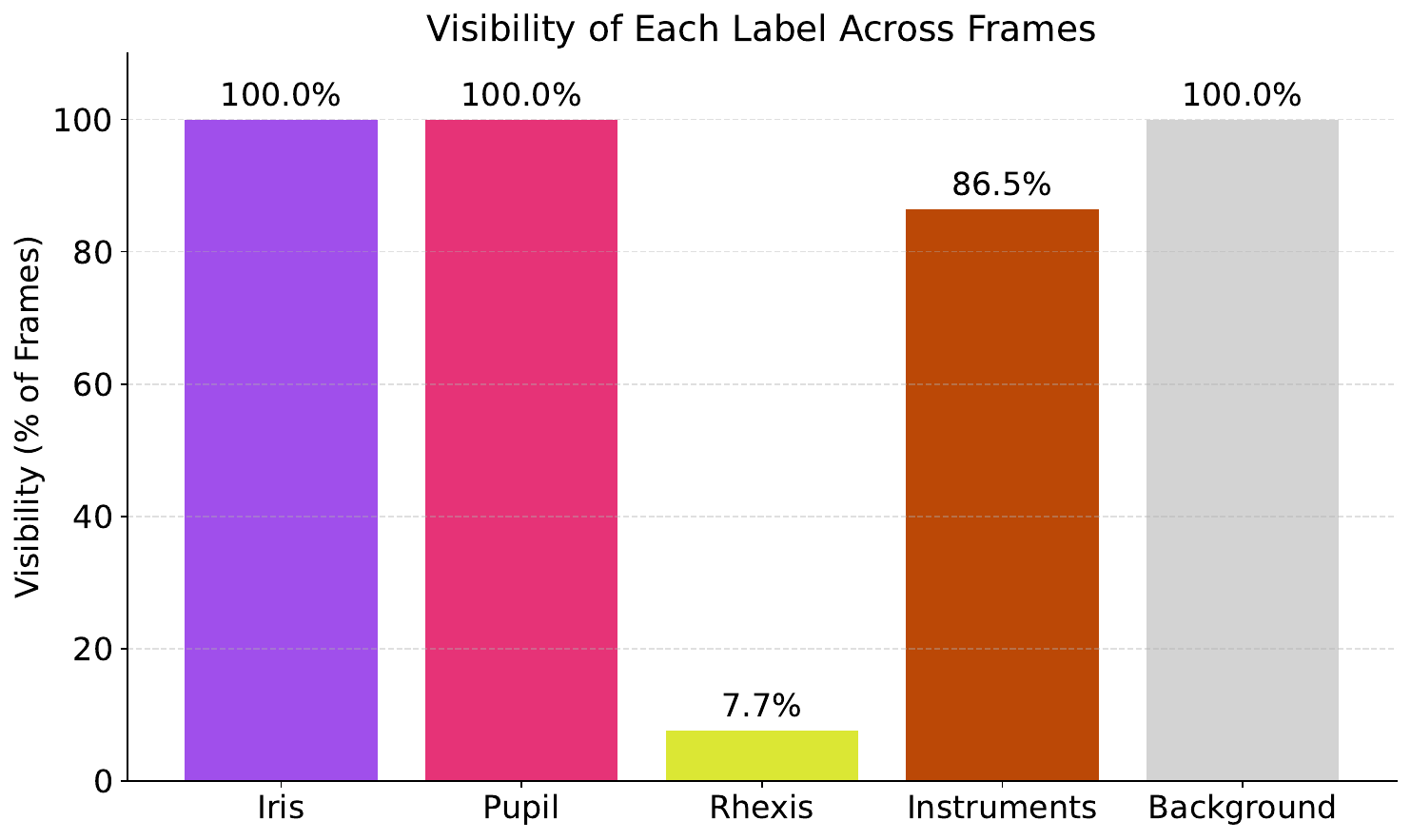}
    \end{subfigure}
    
    \vspace{1.5em}  

    \begin{subfigure}[b]{0.8\columnwidth}
        \centering
        \includegraphics[width=\linewidth]{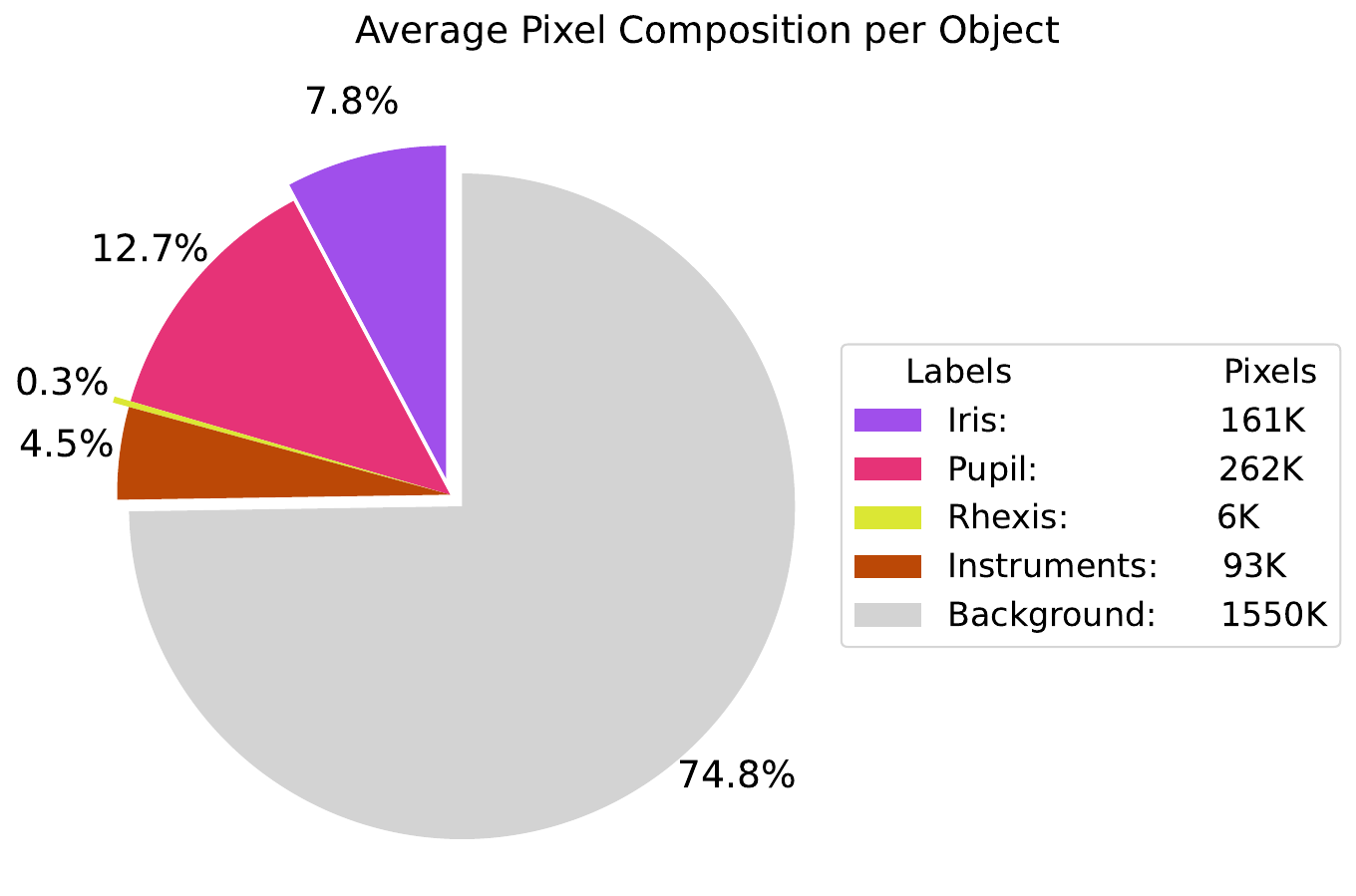}
    \end{subfigure}
    \vspace{0.5em}

    \caption{Comparison of segmentation label visibility and pixel distribution across videos.}
    \Description{Comparison of segmentation label visibility and pixel distribution across videos.}
    \label{fig:segmentation_statistics}
\end{figure}

\section{Dataset}
\label{sec: dataset}

The WetCat dataset comprises 60 cataract surgery videos recorded during 2024–2025, performed by junior to mid-level surgeons using the \textbf{Haag-Streit OSTC} microscope system. Each surgery was conducted under a binocular surgical microscope, providing a magnified, illuminated, and stereoscopic view of the operative field. Surgeons adjusted the microscope’s focus to maintain optimal clarity, while an integrated camera system captured high-resolution video recordings of the complete procedures for comprehensive post-operative analysis and skill evaluation.

Each video in the dataset captures the complete surgical workflow, with an average duration of 794 seconds and a standard deviation of 509 seconds. In addition to the full video recordings, the dataset is comprehensively annotated with surgical phase labels for all cases. Specifically, each video is segmented into four distinct phases: (1) Capsulorhexis (referred to as Rhexis), (2) Phacoemulsification (Phaco.), (3) Idle, and (4) Rest.
Figure~\ref{fig:phase_statistics} summarizes the distribution of phases across all videos and reports the overall percentage of each phase within the dataset. Furthermore, Table~\ref{tbl:statistics_actions} presents normalized phase annotations for a subset of 20 representative videos, providing a detailed overview of phase transitions and relative durations.

\begin{figure}[t!]
    \centering
    \resizebox{1\columnwidth}{!}{%
        \begin{tabular}{ll|ll|ll|ll}
            Iris: & \DP{0.9}{Iris} & 
            Pupil: & \DP{0.9}{Pupil} & 
            Rhexis: & \DP{0.9}{Rhexis} & 
            Instruments: & \DP{0.9}{Instruments}
        \end{tabular}
    }
    \begin{subfigure}[b]{0.45\textwidth}
        \centering
        \includegraphics[width=\linewidth]{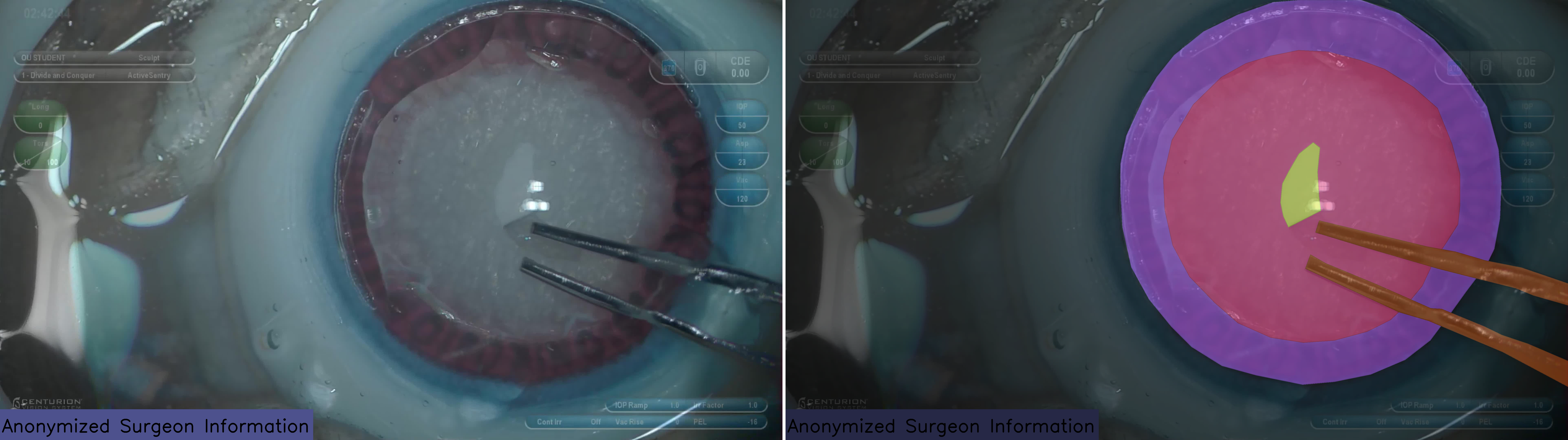}\vspace{-1.5mm} 
        \caption*{Beginning of the Capsularhexis phase.}
    \end{subfigure}
    \begin{subfigure}[b]{0.45\textwidth}
        \centering
        \includegraphics[width=\linewidth]{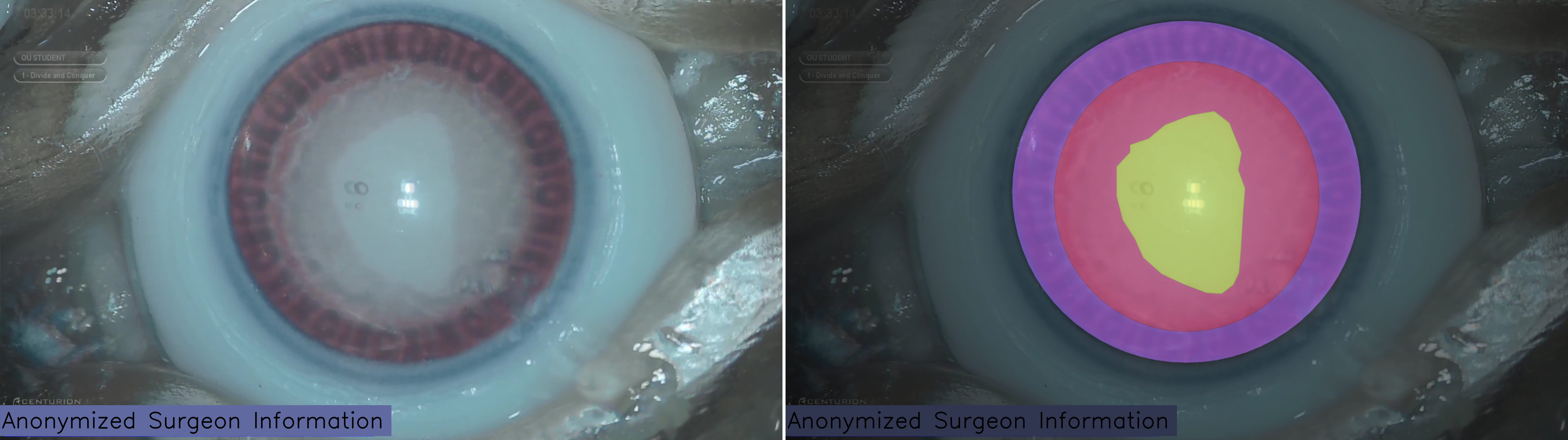}\vspace{-1.5mm} 
        \caption*{End of the Capsularhexis phase.}
    \end{subfigure}
    \begin{subfigure}[b]{0.45\textwidth}
        \centering
        \includegraphics[width=\linewidth]{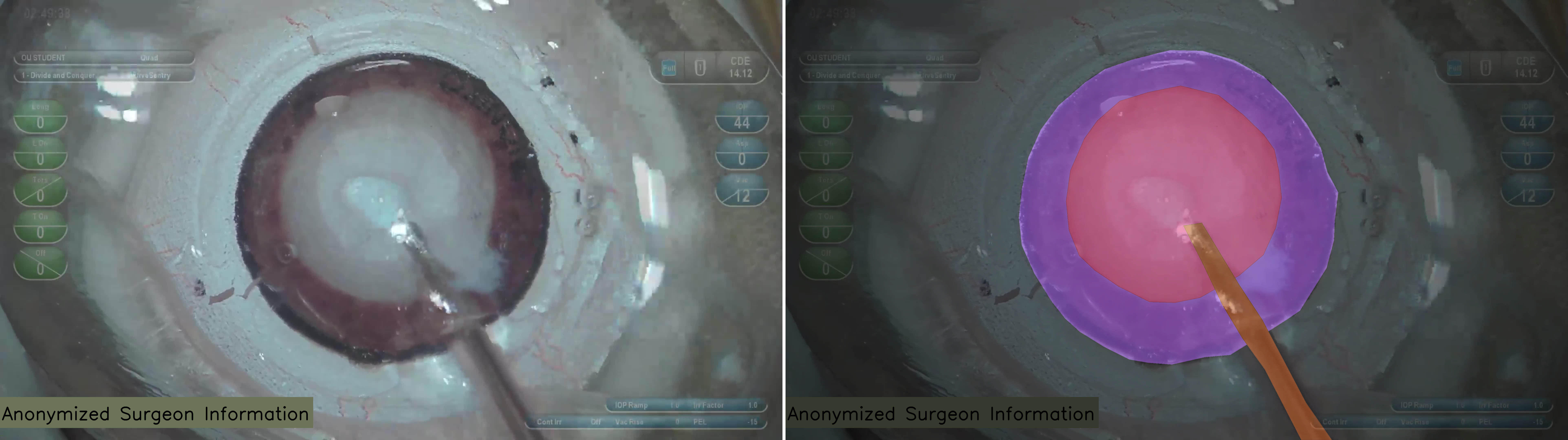}\vspace{-1.5mm} 
        \caption*{Beginning of the Phacoemulsification phase.}
    \end{subfigure}
    \begin{subfigure}[b]{0.45\textwidth}
        \centering
        \includegraphics[width=\linewidth]{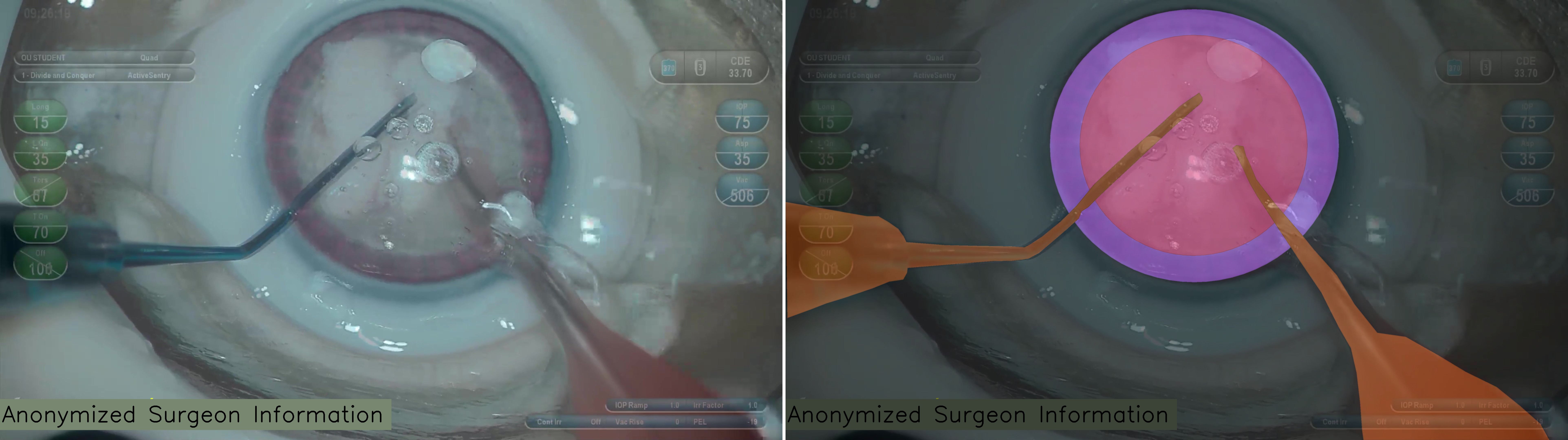}\vspace{-1.5mm} 
        \caption*{End of the Phacoemulsification phase.}
    \end{subfigure}
    \caption{Sample frames from relevant phases in a wet-lab cataract surgery.}
    \Description{Sample frames from relevant phases in a wet-lab cataract surgery.}
    \label{fig:sample-frame-vis}
\end{figure}

In addition to phase annotations, the dataset includes dense pixel-level semantic segmentations for 1,469 selected frames. These segmentations encompass key anatomical structures, including the iris and pupil, as well as surgical instruments and the rhexis region.
Figure~\ref{fig:segmentation_statistics} illustrates the visibility distribution of different labels across frames and depicts the relative pixel-wise percentage for each annotated category within the dataset. Representative frames from the major surgical phases, phacoemulsification and capsulorhexis, along with their corresponding annotations, are visualized in Figure~\ref{fig:sample-frame-vis}.

This comprehensive annotation framework enables detailed analysis of both surgical workflow and scene composition, supporting a wide range of computer vision tasks relevant to automated surgical skill assessment. Table~\ref{tab:dataset_comparison} provides a comparative summary of the WetCat dataset alongside existing datasets focused on phase recognition and semantic segmentation in cataract surgery.

For calibration and scale normalization, anatomical measurements from the OKULO-ONE DIMS artificial eye model are used: the limbus diameter is 11.9 mm, the pupil diameter is 8.5 mm, and the lens diameter is 10 mm. The anterior chamber depth (ACD) is nominally 3.75 mm, while the lens thickness measures approximately 3.87 mm under standard intraocular pressure conditions.

\begin{table}[t!]
\centering
\caption{Visualization of relevant phase annotations for ten representative wetlab videos from our dataset.}
\label{tbl:statistics_actions}
\resizebox{1\columnwidth}{!}{%
\begin{tabular}{m{1cm}m{9cm}m{1cm}m{9cm}}
Case & Phases \\\midrule
1&\DP{0.33}{gray}\DP{0.21}{RedOrange}\DP{0.06}{gray}\DP{0.07}{RedOrange}\DP{0.02}{gray}\DP{0.07}{RedOrange}\DP{0.04}{gray}\DP{0.10}{RedOrange}\DP{0.10}{gray}\DP{0.09}{RedOrange}\DP{0.08}{gray}\DP{0.06}{RedOrange}\DP{0.01}{gray}\DP{0.05}{RedOrange}\DP{0.04}{gray}\DP{0.25}{RedOrange}\DP{0.18}{gray}\DP{0.16}{ForestGreen}\DP{0.17}{gray}\DP{0.78}{ForestGreen}\DP{0.13}{gray}\DP{0.65}{ForestGreen}\DP{0.08}{gray}\DP{0.32}{ForestGreen}\DP{0.09}{gray}\DP{0.03}{RedOrange}\DP{0.18}{gray}\DP{0.12}{ForestGreen}\DP{0.04}{RedOrange}\DP{0.15}{gray}\DP{0.19}{ForestGreen}\DP{0.33}{gray}\DP{0.25}{RoyalBlue}\DP{1.06}{gray}\DP{0.07}{RedOrange}\DP{0.05}{gray}\DP{0.08}{RedOrange}\DP{0.00}{gray}\DP{3.21}{RoyalBlue}\DP{0.11}{gray}\\
2&\DP{0.15}{gray}\DP{0.10}{RedOrange}\DP{0.08}{gray}\DP{0.05}{RedOrange}\DP{0.21}{gray}\DP{0.10}{RedOrange}\DP{0.67}{gray}\DP{0.01}{RedOrange}\DP{0.12}{gray}\DP{0.25}{ForestGreen}\DP{0.09}{gray}\DP{0.28}{ForestGreen}\DP{0.02}{gray}\DP{0.29}{ForestGreen}\DP{0.02}{gray}\DP{0.05}{ForestGreen}\DP{0.04}{gray}\DP{0.13}{ForestGreen}\DP{0.07}{gray}\DP{0.30}{ForestGreen}\DP{0.35}{gray}\DP{0.49}{RoyalBlue}\DP{0.16}{gray}\DP{0.09}{RoyalBlue}\DP{0.20}{gray}\DP{0.13}{RoyalBlue}\DP{0.26}{gray}\DP{4.67}{RoyalBlue}\DP{0.19}{gray}\DP{0.20}{RoyalBlue}\DP{0.21}{gray}\\
3&\DP{0.02}{gray}\DP{0.12}{RedOrange}\DP{0.05}{gray}\DP{0.05}{RedOrange}\DP{0.05}{gray}\DP{0.06}{RedOrange}\DP{0.00}{gray}\DP{0.21}{RedOrange}\DP{0.00}{gray}\DP{0.19}{ForestGreen}\DP{0.00}{gray}\DP{0.08}{RedOrange}\DP{0.00}{gray}\DP{0.49}{ForestGreen}\DP{0.08}{gray}\DP{0.06}{ForestGreen}\DP{0.04}{gray}\DP{0.14}{ForestGreen}\DP{0.04}{gray}\DP{0.05}{RedOrange}\DP{0.05}{gray}\DP{0.03}{ForestGreen}\DP{0.03}{gray}\DP{0.13}{ForestGreen}\DP{0.03}{gray}\DP{0.12}{RedOrange}\DP{0.00}{gray}\DP{0.07}{ForestGreen}\DP{0.04}{gray}\DP{0.07}{ForestGreen}\DP{0.04}{gray}\DP{0.16}{ForestGreen}\DP{0.07}{gray}\DP{0.19}{ForestGreen}\DP{0.09}{gray}\DP{0.31}{RedOrange}\DP{0.05}{gray}\DP{0.33}{ForestGreen}\DP{1.19}{gray}\DP{0.07}{RedOrange}\DP{0.25}{gray}\DP{0.03}{RedOrange}\DP{0.04}{gray}\DP{0.05}{RedOrange}\DP{0.04}{gray}\DP{0.23}{RedOrange}\DP{0.03}{gray}\DP{1.49}{RoyalBlue}\DP{0.21}{gray}\DP{0.14}{RoyalBlue}\DP{0.33}{gray}\DP{1.10}{RoyalBlue}\DP{1.20}{gray}\\
4&\DP{1.16}{gray}\DP{0.24}{RedOrange}\DP{0.00}{gray}\DP{8.18}{RoyalBlue}\DP{0.42}{gray}\\
5&\DP{0.04}{gray}\DP{1.24}{ForestGreen}\DP{0.07}{gray}\DP{0.18}{ForestGreen}\DP{0.05}{gray}\DP{0.25}{RedOrange}\DP{0.02}{gray}\DP{0.33}{RedOrange}\DP{0.04}{gray}\DP{0.08}{RedOrange}\DP{0.15}{gray}\DP{0.09}{RedOrange}\DP{0.11}{gray}\DP{0.09}{RedOrange}\DP{0.24}{gray}\DP{0.02}{RedOrange}\DP{0.07}{gray}\DP{0.07}{RedOrange}\DP{0.16}{gray}\DP{5.88}{RoyalBlue}\DP{0.83}{gray}\\
6&\DP{9.81}{RoyalBlue}\DP{0.19}{gray}\\
7&\DP{0.02}{gray}\DP{0.39}{RedOrange}\DP{0.09}{gray}\DP{0.63}{RedOrange}\DP{0.21}{gray}\DP{0.15}{RedOrange}\DP{0.41}{gray}\DP{0.29}{RedOrange}\DP{0.07}{gray}\DP{1.97}{RoyalBlue}\DP{0.33}{gray}\DP{3.11}{RoyalBlue}\DP{2.33}{gray}\\
8&\DP{0.01}{gray}\DP{0.27}{RedOrange}\DP{0.07}{gray}\DP{0.05}{RedOrange}\DP{0.46}{gray}\DP{6.97}{RoyalBlue}\DP{0.97}{gray}\DP{0.98}{RoyalBlue}\DP{0.22}{gray}\\
9&\DP{0.43}{RedOrange}\DP{0.20}{gray}\DP{0.12}{RedOrange}\DP{0.04}{gray}\DP{2.98}{RoyalBlue}\DP{0.49}{gray}\DP{0.62}{RedOrange}\DP{0.05}{gray}\DP{0.25}{RedOrange}\DP{0.29}{gray}\DP{0.53}{RedOrange}\DP{0.12}{gray}\DP{3.73}{RoyalBlue}\DP{0.14}{gray}\\
10&\DP{0.23}{gray}\DP{0.12}{RedOrange}\DP{0.05}{gray}\DP{0.29}{RedOrange}\DP{0.09}{gray}\DP{0.73}{RedOrange}\DP{0.06}{gray}\DP{0.60}{RedOrange}\DP{0.05}{gray}\DP{0.23}{RedOrange}\DP{0.12}{gray}\DP{0.08}{RedOrange}\DP{0.01}{gray}\DP{0.17}{RedOrange}\DP{0.10}{gray}\DP{0.28}{RedOrange}\DP{0.27}{gray}\DP{0.97}{RoyalBlue}\DP{2.64}{gray}\DP{2.23}{RoyalBlue}\DP{0.69}{gray}\\
11&\DP{0.07}{gray}\DP{0.02}{RedOrange}\DP{0.24}{gray}\DP{0.39}{RedOrange}\DP{0.20}{gray}\DP{0.22}{RedOrange}\DP{0.33}{gray}\DP{0.07}{RedOrange}\DP{0.09}{gray}\DP{0.23}{RedOrange}\DP{0.00}{gray}\DP{0.03}{RedOrange}\DP{0.11}{gray}\DP{0.02}{RedOrange}\DP{0.77}{gray}\DP{6.33}{RoyalBlue}\DP{0.88}{gray}\\
12&\DP{0.11}{gray}\DP{4.10}{RoyalBlue}\DP{1.63}{gray}\DP{4.01}{RoyalBlue}\DP{0.16}{gray}\\
13&\DP{3.04}{RoyalBlue}\DP{0.89}{gray}\DP{0.19}{RedOrange}\DP{3.06}{gray}\DP{1.85}{RoyalBlue}\DP{0.47}{gray}\DP{0.18}{RedOrange}\DP{0.33}{gray}\\
14&\DP{0.38}{RedOrange}\DP{0.07}{gray}\DP{0.46}{RedOrange}\DP{1.38}{gray}\DP{0.21}{RedOrange}\DP{0.18}{gray}\DP{3.80}{RoyalBlue}\DP{0.61}{gray}\DP{2.62}{RoyalBlue}\DP{0.29}{gray}\\
15&\DP{0.16}{gray}\DP{0.09}{RedOrange}\DP{0.09}{gray}\DP{0.27}{RedOrange}\DP{0.27}{RedOrange}\DP{0.08}{gray}\DP{0.63}{RedOrange}\DP{0.15}{gray}\DP{0.14}{RedOrange}\DP{0.04}{gray}\DP{1.18}{RedOrange}\DP{0.07}{gray}\DP{0.56}{RedOrange}\DP{0.03}{gray}\DP{0.55}{ForestGreen}\DP{0.05}{gray}\DP{0.58}{RedOrange}\DP{0.02}{gray}\DP{1.00}{ForestGreen}\DP{0.13}{gray}\DP{0.13}{RedOrange}\DP{0.05}{gray}\DP{1.03}{RedOrange}\DP{0.16}{gray}\DP{0.13}{RedOrange}\DP{0.26}{gray}\DP{0.26}{RedOrange}\DP{0.39}{gray}\DP{0.07}{RedOrange}\DP{1.43}{gray}\\
16&\DP{2.63}{ForestGreen}\DP{0.13}{gray}\DP{0.19}{ForestGreen}\DP{0.07}{gray}\DP{0.17}{ForestGreen}\DP{0.00}{gray}\DP{6.79}{RoyalBlue}\DP{0.00}{gray}\\
17&\DP{1.60}{ForestGreen}\DP{0.03}{gray}\DP{0.85}{ForestGreen}\DP{0.05}{gray}\DP{7.42}{RoyalBlue}\DP{0.04}{gray}\\
18&\DP{0.01}{gray}\DP{1.80}{ForestGreen}\DP{0.10}{gray}\DP{0.62}{ForestGreen}\DP{0.18}{gray}\DP{7.18}{RoyalBlue}\DP{0.12}{gray}\\
19&\DP{0.21}{ForestGreen}\DP{0.21}{gray}\DP{1.56}{ForestGreen}\DP{0.02}{gray}\DP{0.15}{ForestGreen}\DP{0.02}{gray}\DP{7.57}{RoyalBlue}\DP{0.25}{gray}\\
20&\DP{0.01}{gray}\DP{2.99}{ForestGreen}\DP{0.12}{gray}\DP{4.70}{RoyalBlue}\DP{0.27}{gray}\DP{1.90}{RoyalBlue}\DP{0.00}{gray}\\

\specialrule{.12em}{.05em}{.05em}
\end{tabular}}
\resizebox{1\columnwidth}{!}{%
\begin{tabular}{m{1.5cm}m{10cm}}
Colormap & 
Rhexis \DP{0.9}{ForestGreen},
Phaco. \DP{0.9}{RoyalBlue},
Rest \DP{0.9}{RedOrange},
Idle \DP{0.9}{gray}
\\
\specialrule{.12em}{.05em}{.05em}
\end{tabular}
}
\end{table}

\begin{table}[b!]
\centering
\caption{Comparison of annotated subsets in the WetCat dataset with existing datasets for semantic segmentation and phase recognition in cataract surgery.}
\label{tab:dataset_comparison}
\resizebox{\columnwidth}{!}{%
\begin{tabular}{>{\raggedright\arraybackslash}p{2.8cm}
  >{\centering\arraybackslash}m{1.4cm}
  >{\centering\arraybackslash}m{1.4cm}
  >{\centering\arraybackslash}m{1.4cm}
  >{\centering\arraybackslash}m{1.4cm}
  >{\centering\arraybackslash}m{1.4cm}}
\toprule
\textbf{Attribute} & \textbf{Type} & \textbf{CaDIS}~\cite{CaDIS} & \textbf{CatRel}~\cite{LocalPhase} & \textbf{Cataract-1K}~\cite{Cataract-1K_dataset} & \textbf{WetCat} \\
\midrule
\textbf{Acquisition Period} & & 2015 & 2017--2018 & 2021--2023 & 2024--2025 \\
\textbf{Phase Annotations} & \# Videos & \xmark & 22 & 56 & 60 \\
\textbf{Segment Annotations} & \# Frames & 4,670 & \xmark & 2,256 & 1,469 \\
\textbf{Phase Classes} & & \xmark & 5 & 13 & 4 \\
\textbf{Segment Classes} & Anatomy & \cmark & N/A & \cmark & \cmark \\
                                 & Instruments & \cmark & N/A & \cmark & \cmark \\
                                 & Rhexis & \xmark & N/A & \xmark & \cmark \\
\textbf{Resolution} & & $960 \times 540$ & $224 \times 224$ & $1024 \times 768$ & $1920 \times 1080$ \\
\textbf{Frame Rate (fps)} & & N/A & 25 & 30 & 30 \\
\bottomrule
\end{tabular}
 }
\end{table}

\paragraph{Usage Notes}
The datasets are licensed under \href{https://creativecommons.org/licenses/by-nc-sa/4.0/}{CC BY-NC-SA 4.0}. We provide all code for annotation processing, as well as the training IDs for four-fold validation and usage instructions in the GitHub repository of the paper (\url{https://github.com/Negin-Ghamsarian/WetCat}).

\section{Technical Validation}
\label{sec:validation}

In this section, we validate the quality and consistency of our multi-task annotations by training a set of state-of-the-art neural network architectures, each tailored to its respective task. The performance of these models is thoroughly assessed using task-appropriate evaluation metrics, providing a robust measure of the accuracy and reliability of our annotations.

\subsection{Experimental Settings for Phase Recognition}

We evaluate three types of neural network architectures for phase recognition. 
First, we adopt a combined CNN-RNN framework, where the CNN backbone is VGG16, and the RNN component is configured with four variants: GRU, LSTM, BiGRU, and BiLSTM, following the architectures proposed in \cite{LocalPhase}.
Second, we implement a vision transformer model that integrates VGG16 as the backbone with a lightweight transformer head comprising 4.30 M parameters, as described in \cite{nasirihaghighi2024event}.
Third, we employ a 3D convolutional residual network (ResNet3D-18) following \cite{ResNet3D}.
All backbone networks are pre-trained on ImageNet \cite{ImageNet}, except for ResNet3D, which is trained from scratch.

For phase recognition, we merge the "Idle" and "Rest" phases into a single class, resulting in a three-class segmentation task.
Phase recognition is conducted on three-second video clips by randomly sampling 10 frames from 90 frames within each clip to construct the input sequences, as per the protocols in \cite{LocalPhase,10178763}.
We evaluate the performance using accuracy, macro recall, and macro F1 score as the primary metrics.

\begin{table}[t!]
\renewcommand{\arraystretch}{1}
\caption{Specifications of the proposed and alternative approaches.}

\label{tab:alternatives}
\centering
\resizebox{1\columnwidth}{!}{%
\begin{tabular}{>{\raggedright\arraybackslash}p{2.5cm}
  >{\centering\arraybackslash}m{1.7cm}
  >{\centering\arraybackslash}m{1.7cm}
  >{\centering\arraybackslash}m{2cm}}
\specialrule{.12em}{.05em}{.05em}
Model & Backbone & Params. &  Target \\\specialrule{.12em}{.05em}{.05em}
Adapt-Net \cite{LensID} & VGG16 & 24.69 M &  Surgical Videos  \\
UNet$++$~\cite{UNet++}&VGG16&24.24 M&  Medical Images  \\
ReCal-Net \cite{ReCal-Net}& VGG16 & 22.93 M &  Surgical Videos \\
CPFNet \cite{CPFNet}& VGG16  & 39.17 M &  Medical Images \\
CE-Net \cite{CE-Net}& ResNet34& 29.90 M&  Medical Images  \\
DeepLabV3+ \cite{DeepLabV3}& ResNet50& 26.68 M&Natural Images \\
UPerNet \cite{UPerNet}& ResNet50& 51.26 M&Natural Images \\
U-Net+ \cite{U-Net} \footnote{Note that UNet+ is an improved version of UNet, where we use VGG16 as the backbone network and double convolutional blocks (two consecutive convolutions followed by batch normalization and ReLU layers) as decoder modules.}& VGG16 &22.55 M& Medical Images \\
SAM \cite{kirillov2023segment}& ViT-B & 91 M & Natural Images \\
\specialrule{.12em}{.05em}{0.05em}
\end{tabular}
}
\end{table}

\subsection{Experimental Settings for Semantic Segmentation}

We conduct a series of experiments to assess the reliability of our pixel-level annotations, benchmarking against multiple state-of-the-art methods designed for natural images, medical images, and surgical videos. Detailed descriptions of these baseline approaches are provided in Table~\ref{tab:alternatives}.

Given the substantial number of parameters in the Segment Anything Model (SAM) and the considerable computational resources required for full training, we adopt an efficient fine-tuning strategy.
Specifically, we freeze both the vision encoder and the prompt encoder, and investigate two training configurations: (1) training only the mask decoder, resulting in 4,058,340 trainable parameters, and (2) applying Low-Rank Adaptation (LoRA) \cite{hu2022lora} to the linear and convolutional (Conv2D) layers of the vision encoder, with a rank of 16 ($r=16$), scaling factor $\alpha=32$, and a dropout rate of 0.1, increasing the trainable parameters to 6,647,012. Unless otherwise specified, SAM is fine-tuned using grid prompts as input. 

We evaluate the baselines by computing the average Dice coefficient and the average Intersection over Union (IoU).

\subsection{Training Settings}
All neural networks are initialized with ImageNet pre-trained weights \cite{ImageNet} for their respective backbones, except for ResNet3D and SAM.

For phase recognition models, we use a batch size of 16 and input frame dimensions of \(224 \times 224\) pixels.
For semantic segmentation models, we use a batch size of four and resize all input images to \(512 \times 512\) pixels.

The initial learning rate is set to 0.05 for ResNet3D and 0.005 for the CNN-RNN and transformer-based models.
For semantic segmentation, the initial learning rate is set to 0.001 for convolutional networks and 0.0001 for SAM.
For networks with pre-trained backbones, the backbone's learning rate is set to one-tenth of the main learning rate.
The learning rate is progressively reduced during training according to a polynomial decay schedule, as described in \cite{ghamsarian2024cataract}.
To enhance model robustness and promote generalization, we apply a range of data augmentations during training for both tasks, including random resized cropping (scale variation between 0.8 and 1.2), random rotations (up to \(\pm 30^{\circ}\)), color jittering (brightness, contrast, and saturation adjustments up to 20\%), Gaussian blurring, random sharpening, and random conversion to grayscale with a probability of 0.2.

Phase recognition models are optimized using the cross-entropy loss function.
Semantic segmentation models are trained using the \textit{cross-entropy log-dice} loss, defined as:
\begin{equation*}
\begin{split}
    \mathcal{L} = (\lambda)\times CE(\mathcal{X}_{true}(i,j),\mathcal{X}_{pred}(i,j)) \\
    -(1-\lambda)\times \left( \log \frac{2\sum \mathcal{X}_{true}\odot \mathcal{X}_{pred}+\sigma}{\sum \mathcal{X}_{true} + \sum \mathcal{X}_{pred}+ \sigma} \right)
\end{split}
\label{eq:loss}
\end{equation*}

\noindent Here, $\mathcal{X}_{true}$ represents the ground truth mask, and $\mathcal{X}_{pred}$ denotes the predicted mask, constrained such that $0 \leq \mathcal{X}_{pred}(i,j) \leq 1$.
The weighting parameter $\lambda \in [0,1]$ is set to 0.8 in our experiments.
The symbol $\odot$ indicates the Hadamard product (element-wise multiplication), and $\sigma$ is the Laplacian smoothing constant set to 1 to prevent numerical instability and mitigate overfitting.

\begin{table}[t]
\caption{Phase recognition performance in wetlab cataract surgery.}
    \centering
    \resizebox{1\columnwidth}{!}{%
\begin{tabular}{>{\raggedright\arraybackslash}p{2.5cm}
  >{\centering\arraybackslash}m{1.5cm}
  >{\centering\arraybackslash}m{1.5cm}
  >{\centering\arraybackslash}m{1.5cm}}
\toprule

Network & Accuracy \% & Recall \% & F1-Score \% \\

\midrule
ResNet3D & 80.77 \tiny{\(\pm 1.41\)} & 71.17 \tiny{\(\pm 3.08\)} & 80.36 \tiny{\(\pm 2.59\)} \\
VGG-LSTM & 85.19 \tiny{\(\pm 4.65\)} & 84.00 \tiny{\(\pm 4.47\)} & 84.85 \tiny{\(\pm 4.45\)} \\
VGG-GRU & 83.90 \tiny{\(\pm 3.40\)} & 83.42 \tiny{\(\pm 4.14\)} & 84.58 \tiny{\(\pm 4.26\)} \\
VGG-BiLSTM &84.84 \tiny{\(\pm 4.72\)} & 85.02 \tiny{\(\pm 4.37\)} & 85.51 \tiny{\(\pm 4.55\)} \\
VGG-BiGRU & 82.80 \tiny{\(\pm 5.13\)} & 82.71 \tiny{\(\pm 5.65\)} & 84.02 \tiny{\(\pm 5.26\)} \\
VGG-Transformer & 79.75 \tiny{\(\pm 1.45\)} & 77.98 \tiny{\(\pm 2.77\)} & 80.79 \tiny{\(\pm 1.72\)} \\

\bottomrule
\end{tabular}
}
    
    \label{tab:phase-recognition}
\end{table}

\begin{table*}[tb]
\caption{Quantitative evaluations of anatomy and instruments segmentation performance for neural network architectures listed in Table \ref{tab:alternatives}.}
    \centering
    \resizebox{1\textwidth}{!}{%
\begin{tabular}{ll*{8}{>{\centering\arraybackslash}m{1.4cm}}}
\toprule
&& \multicolumn{3}{c}{Anatomy \footnotesize{(IoU \%)}} & \multicolumn{3}{c}{Anatomy \footnotesize{(Dice \%)}} & \multicolumn{2}{c}{\multirow{2}{*}{Instruments \footnotesize{(IoU/Dice \%)}}}\\
\cmidrule(lr){3-5}\cmidrule(lr){6-8}
Backbone & Network & Iris & Pupil & Avg. & Iris & Pupil & Avg. & &\\
\midrule
\multirow{5}{*}{VGG16}& UNet+ & 75.55 \tiny{\(\pm 5.42\)} & 81.19 \tiny{\(\pm 6.37\)} & 78.37 & 85.96 \tiny{\(\pm 3.64\)} & 89.48 \tiny{\(\pm 3.95\)} & 87.72 & 60.00 \tiny{\(\pm 2.67\)} & 69.48 \tiny{\(\pm 2.63\)} \\
& CPFNet & 77.39 \tiny{\(\pm 5.79\)} & 82.21 \tiny{\(\pm 11.58\)} & 79.80 & 87.13 \tiny{\(\pm 3.81\)} & 89.76 \tiny{\(\pm 7.49\)} & 88.45 & 65.92 \tiny{\(\pm 3.13\)} & 74.72 \tiny{\(\pm 2.79\)} \\
& UNetPP & 75.42 \tiny{\(\pm 6.15\)} & 80.47 \tiny{\(\pm 8.13\)} & 77.94 & 85.84 \tiny{\(\pm 4.13\)} & 88.95 \tiny{\(\pm 5.17\)} & 87.39 & 60.55 \tiny{\(\pm 3.04\)} & 70.00 \tiny{\(\pm 2.86\)} \\
& AdaptNet & 72.73 \tiny{\(\pm 4.88\)} & 85.86 \tiny{\(\pm 10.18\)} & 79.29 & 84.12 \tiny{\(\pm 3.38\)} & 92.04 \tiny{\(\pm 6.29\)} & 88.08 & 65.25 \tiny{\(\pm 3.27\)} & 73.66 \tiny{\(\pm 3.16\)} \\
& ReCal-Net & 77.31 \tiny{\(\pm 5.95\)} & 75.52 \tiny{\(\pm 14.78\)} & 76.42 & 87.08 \tiny{\(\pm 3.92\)} & 85.17 \tiny{\(\pm 10.48\)} & 86.13 & 61.67 \tiny{\(\pm 2.23\)} & 70.55 \tiny{\(\pm 1.84\)} \\\midrule
\multirow{3}{*}{ResNet34}& CENet & 77.08 \tiny{\(\pm 9.37\)} & 83.28 \tiny{\(\pm 11.59\)} & 80.18 & 86.72 \tiny{\(\pm 6.36\)} & 90.41 \tiny{\(\pm 7.32\)} & 88.57 & 48.23 \tiny{\(\pm 22.20\)} & 54.82 \tiny{\(\pm 25.94\)} \\
& AdaptNet & 74.50 \tiny{\(\pm 5.68\)} & 84.54 \tiny{\(\pm 12.20\)} & 79.52 & 85.27 \tiny{\(\pm 3.83\)} & 91.11 \tiny{\(\pm 7.69\)} & 88.19 & 68.16 \tiny{\(\pm 2.37\)} & 76.28 \tiny{\(\pm 2.11\)} \\
& ReCal-Net & 77.90 \tiny{\(\pm 5.87\)} & 82.79 \tiny{\(\pm 16.07\)} & 80.34 & 87.45 \tiny{\(\pm 3.84\)} & 89.65 \tiny{\(\pm 10.66\)} & 88.55 & 64.45 \tiny{\(\pm 1.86\)} & 72.55 \tiny{\(\pm 2.33\)} \\\midrule
\multirow{2}{*}{ResNet50}& UPerNet & 80.41 \tiny{\(\pm 5.37\)} & 88.63 \tiny{\(\pm 6.41\)} & 84.52 & 89.04 \tiny{\(\pm 3.40\)} & 93.85 \tiny{\(\pm 3.68\)} & 91.44 & 70.81 \tiny{\(\pm 2.72\)} & 78.83 \tiny{\(\pm 2.31\)} \\
& DeepLabV3+ & 79.97 \tiny{\(\pm 6.02\)} & 88.66 \tiny{\(\pm 7.49\)} & 84.32 & 88.74 \tiny{\(\pm 3.86\)} & 93.82 \tiny{\(\pm 4.32\)} & 91.28 & 70.79 \tiny{\(\pm 2.77\)} & 79.01 \tiny{\(\pm 2.37\)} \\\midrule
\multirow{2}{*}{ViT-B}& SAM & 75.53 \tiny{\(\pm 5.00\)} & 91.20 \tiny{\(\pm 1.88\)} & 83.36 & 84.51 \tiny{\(\pm 4.84\)} & 94.97 \tiny{\(\pm 1.34\)} & 89.74 & 62.22 \tiny{\(\pm 2.81\)} & 74.67 \tiny{\(\pm 2.71\)} \\
& SAM-LoRA & 80.41 \tiny{\(\pm 3.02\)} & 91.63 \tiny{\(\pm 2.50\)} & 86.02 & 88.25 \tiny{\(\pm 2.48\)} & 94.93 \tiny{\(\pm 1.95\)} & 91.59 & 69.79 \tiny{\(\pm 2.35\)} & 80.62 \tiny{\(\pm 2.17\)} \\
\bottomrule
\end{tabular}
}
    
    \label{tab:quantitative-anatomy-inst}
\end{table*}

\begin{table}[tb]
\caption{Quantitative evaluations of rhexis segmentation performance for neural network architectures listed in Table \ref{tab:alternatives}.}
    \centering
    \resizebox{1\columnwidth}{!}{%
\begin{tabular}{>{\raggedright\arraybackslash}p{1.2cm}
  >{\raggedright\arraybackslash}p{2.1cm}
  >{\centering\arraybackslash}m{2cm}
  >{\centering\arraybackslash}m{2cm}}
\toprule

Backbone & Network & IoU \% & Dice \% \\

\midrule
\multirow{5}{*}{VGG16}& UNet+ & \NP{44.05} \tiny{\(\pm 9.85\)} & \NP{55.90} \tiny{\(\pm 10.31\)} \\
& CPFNet & \NP{49.89} \tiny{\(\pm 15.92\)} & \NP{59.21} \tiny{\(\pm 16.54\)} \\
& UNetPP & \NP{44.42} \tiny{\(\pm 17.20\)} & \NP{55.33} \tiny{\(\pm 18.23\)} \\
& AdaptNet & \NP{45.17} \tiny{\(\pm 18.69\)} & \NP{54.52} \tiny{\(\pm 18.51\)} \\
& ReCal-Net & \NP{47.14} \tiny{\(\pm 19.96\)} & \NP{56.15} \tiny{\(\pm 19.89\)} \\\midrule
\multirow{3}{*}{ResNet34}& CENet & \NP{14.53} \tiny{\(\pm 13.53\)} & \NP{16.68} \tiny{\(\pm 16.86\)} \\
& AdaptNet & \NP{51.38} \tiny{\(\pm 17.13\)} & \NP{60.27} \tiny{\(\pm 17.18\)} \\
& ReCal-Net & \NP{52.98} \tiny{\(\pm 19.68\)} & \NP{60.27} \tiny{\(\pm 19.54\)} \\\midrule
\multirow{2}{*}{ResNet50}& UPerNet & \NP{68.83} \tiny{\(\pm 10.35\)} & \NP{77.47} \tiny{\(\pm 9.71\)} \\
& DeepLabV3+ & \NP{65.30} \tiny{\(\pm 11.76\)} & \NP{74.11} \tiny{\(\pm 11.33\)} \\\midrule
\multirow{4}{*}{ViT-B}& SAM & \NP{64.60} \tiny{\(\pm 4.31\)} & \NP{75.14} \tiny{\(\pm 4.88\)} \\
& SAM (bb) & \NP{66.88} \tiny{\(\pm 10.03\)} & \NP{77.55} \tiny{\(\pm 9.41\)} \\
& SAM-LoRA & \NP{74.20} \tiny{\(\pm 3.80\)} & \NP{81.90} \tiny{\(\pm 3.97\)} \\
& SAM-LoRA (bb) & \NP{77.57} \tiny{\(\pm 4.33\)} & \NP{85.79} \tiny{\(\pm 2.97\)} \\
\bottomrule
\end{tabular}
}
    
    \label{tab:quantitative-rhexis}
\end{table}

\begin{figure}[tb]
    \centering
    \begin{subfigure}[t]{0.49\columnwidth}
    \centering
        \includegraphics[width=0.99\columnwidth]{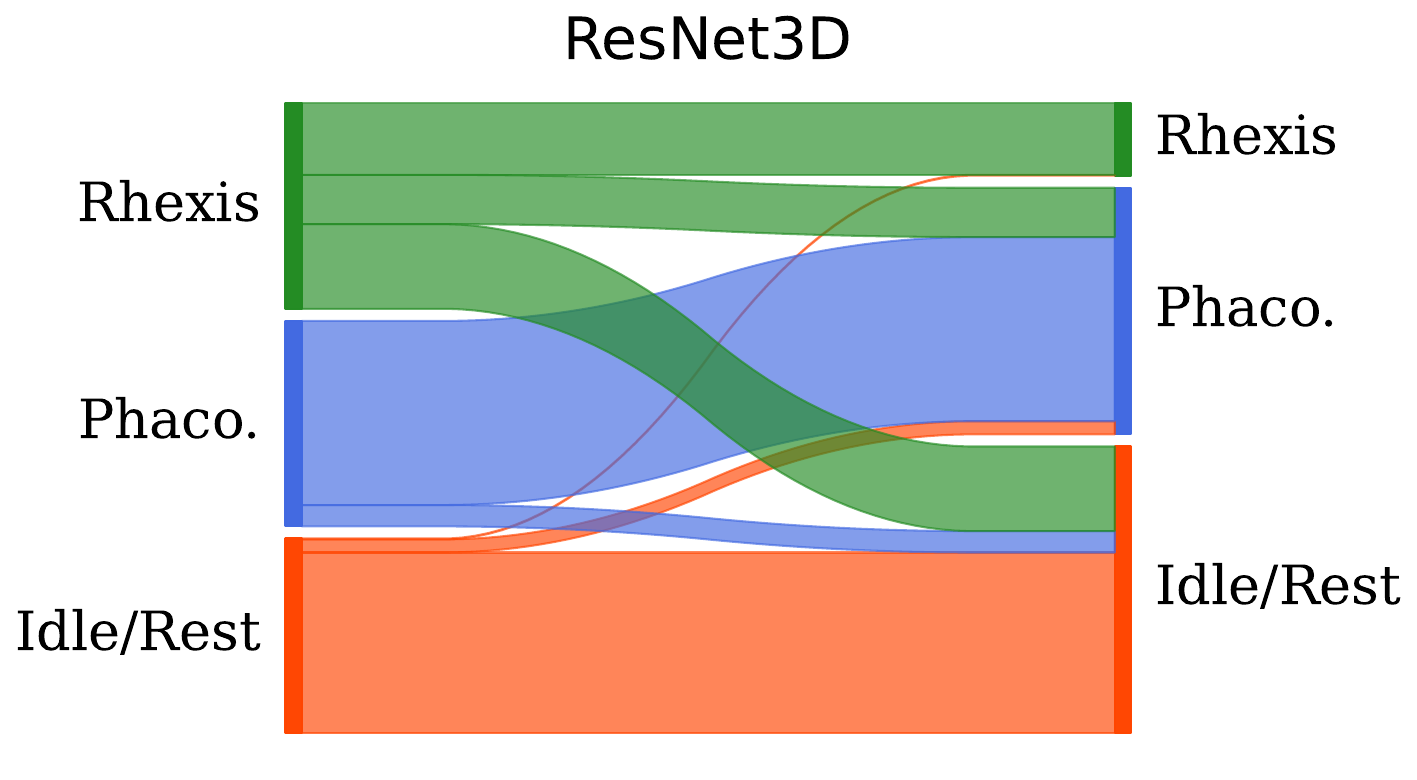} 
    \end{subfigure}
    \begin{subfigure}[t]{0.49\columnwidth}
    \centering
        \includegraphics[width=0.99\columnwidth]{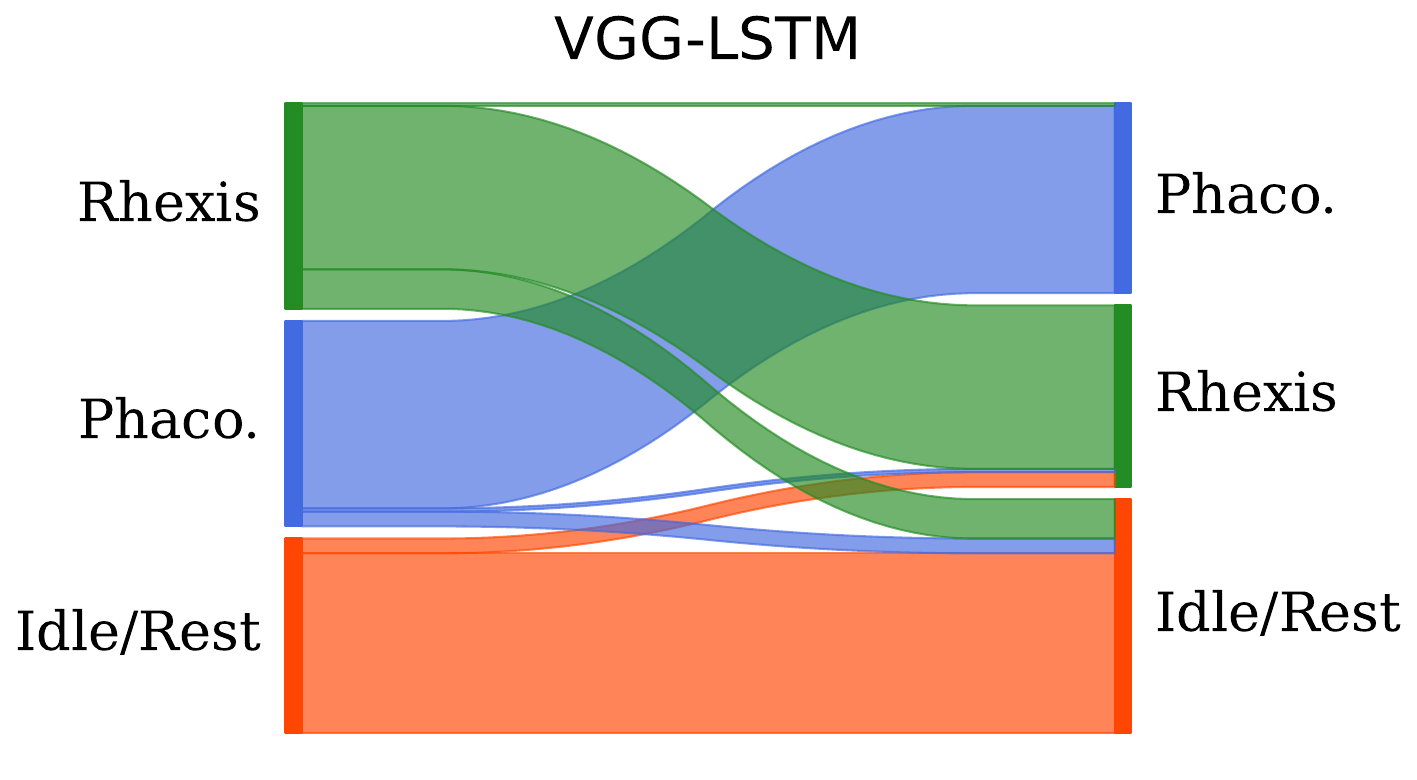}  
    \end{subfigure}
    \\\vspace{1em}
    \begin{subfigure}[t]{0.49\columnwidth}
    \centering
        \includegraphics[width=0.99\columnwidth]{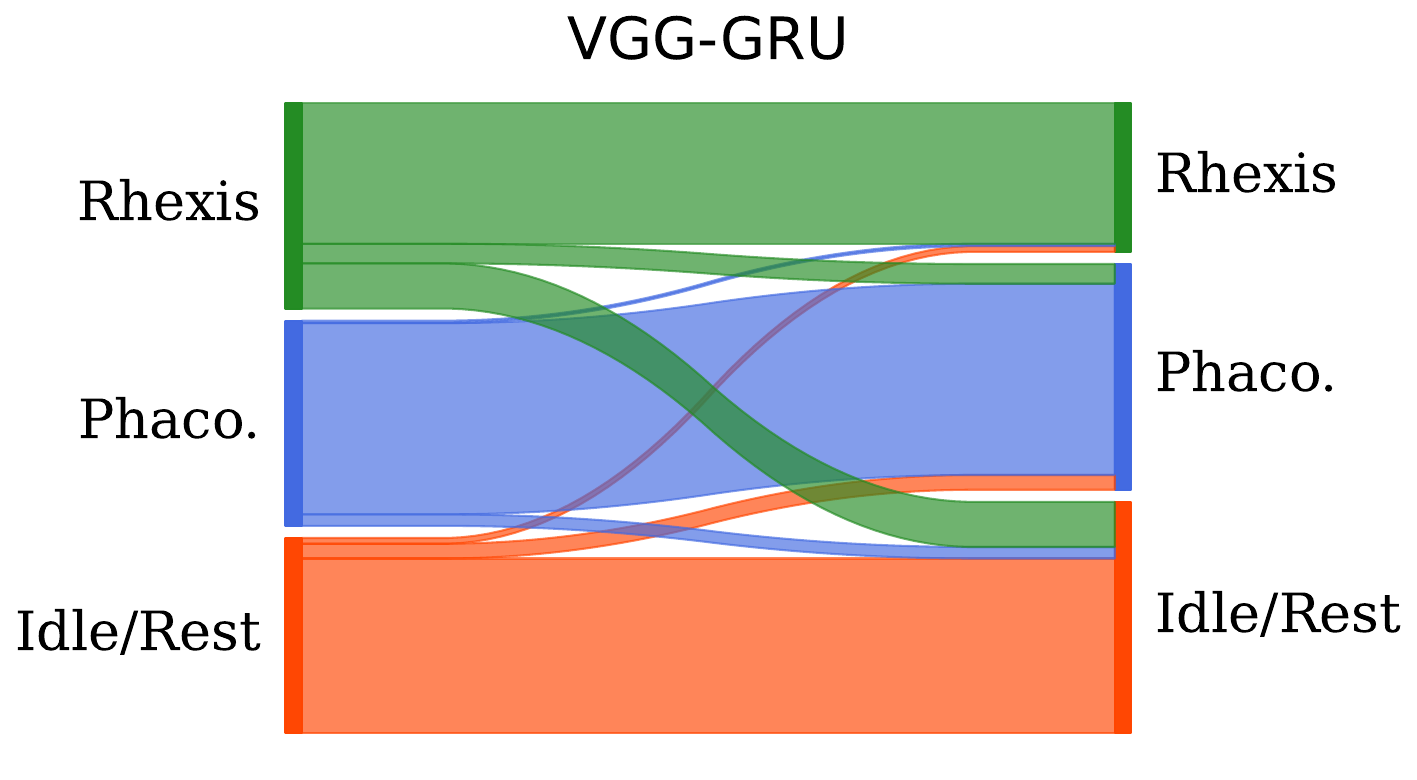} 
    \end{subfigure}
    \begin{subfigure}[t]{0.49\columnwidth}
    \centering
        \includegraphics[width=0.99\columnwidth]{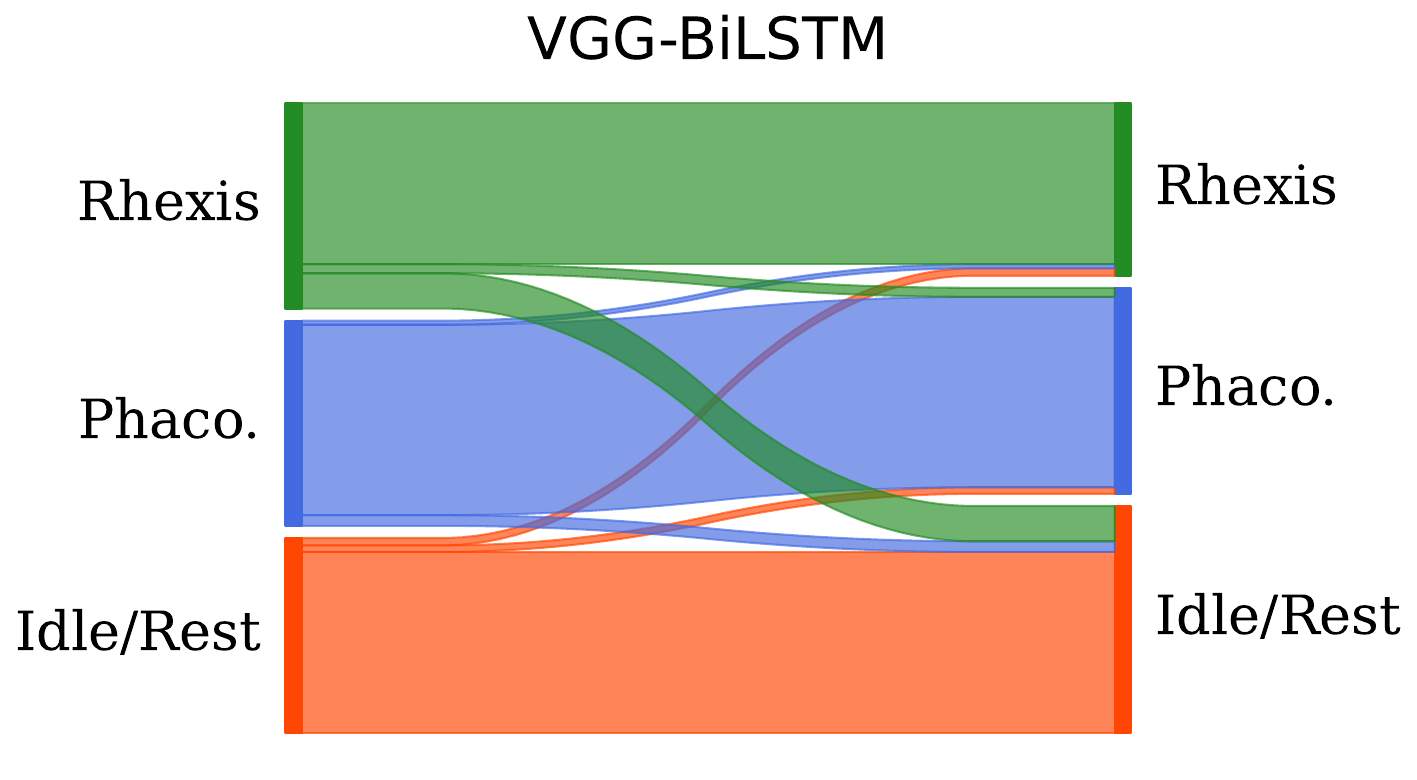}  
    \end{subfigure}
    \\\vspace{1em}
    \begin{subfigure}[t]{0.49\columnwidth}
    \centering
        \includegraphics[width=0.99\columnwidth]{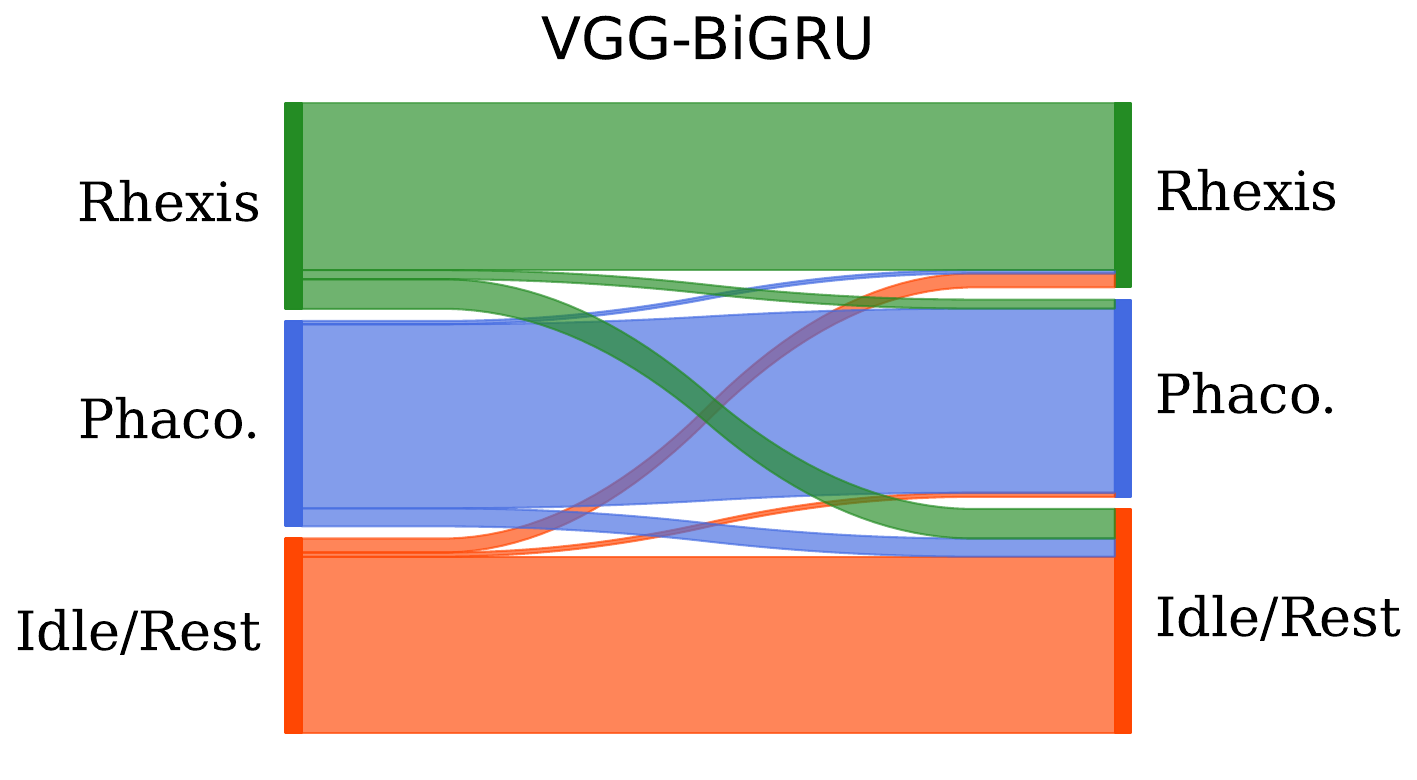} 
    \end{subfigure}
    \begin{subfigure}[t]{0.49\columnwidth}
    \centering
        \includegraphics[width=0.99\columnwidth]{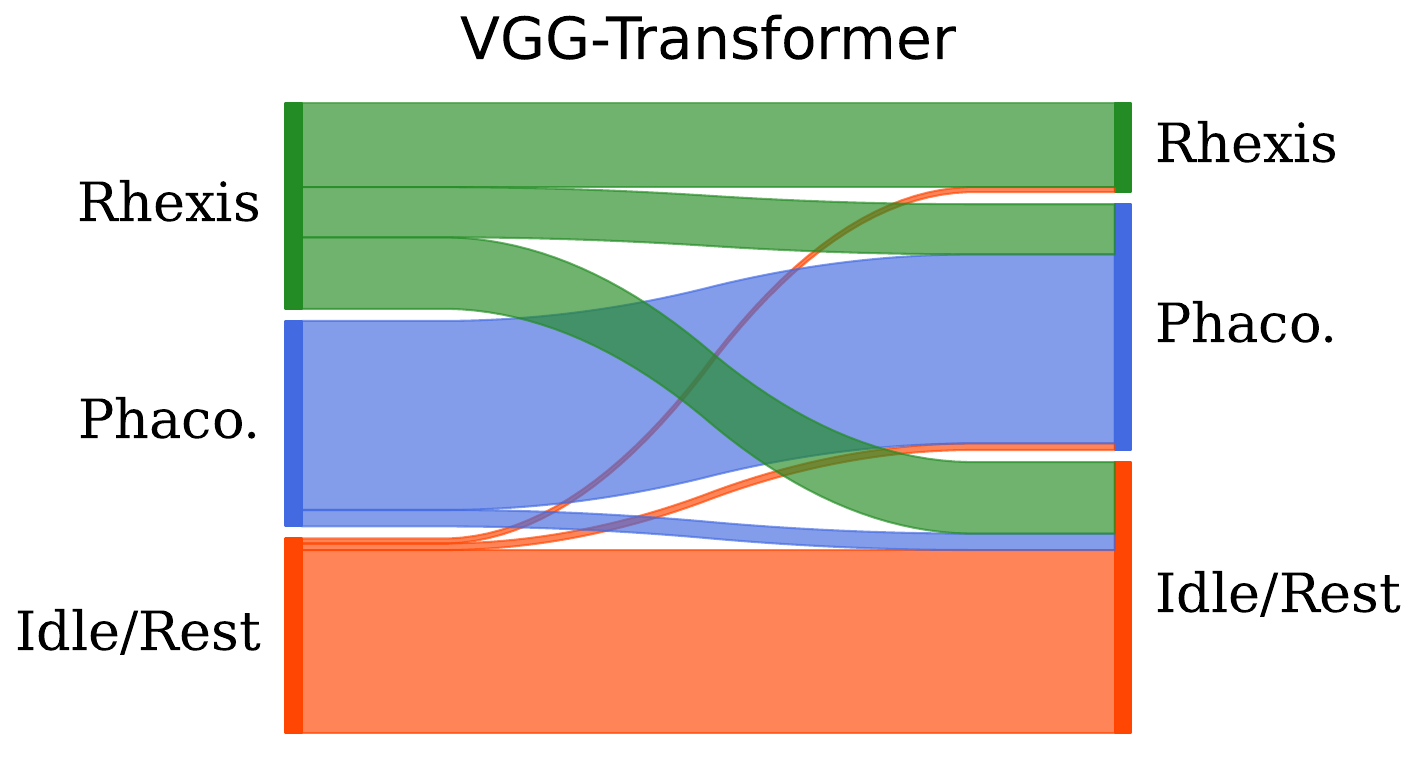}  
    \end{subfigure}
    \caption{Sankey diagrams of confusion matrices corresponding to different phase recognition networks.}
    \Description{Sankey diagrams of confusion matrices corresponding to different phase recognition networks.}
    \label{fig:phases-sankey}
\end{figure}

\subsection{Experimental Results}
Table~\ref{tab:phase-recognition} presents the phase recognition performance of various network architectures, averaged over four cross-validation folds. The results demonstrate strong and consistent performance across different network configurations. In particular, the integration of recurrent layers notably enhances detection accuracy and F1-score across all settings. 
Figure~\ref{fig:phases-sankey} visualizes the confusion matrices corresponding to the evaluated network architectures on a common test fold using Sankey diagrams. The figures reveal that misclassifications predominantly occur between the rhexis and idle/rest phases, with a significant portion of errors involving the misclassification of rhexis as idle/rest. However, these misclassification rates are substantially reduced when using BiGRU and BiLSTM layers, highlighting the ability of bidirectional recurrent structures to better separate visually similar classes by learning more discriminative spatio-temporal representations.

Table~\ref{tab:quantitative-anatomy-inst} reports the quantitative performance of several neural network architectures on anatomical structure and instrument segmentation tasks. The results indicate that segmenting anatomical structures is generally less challenging than segmenting surgical instruments across all evaluated models. Among anatomical categories, pupil segmentation achieves the highest performance, likely due to its well-defined contours and clear boundaries, whereas iris segmentation shows comparatively lower performance, attributed to its less distinct edges. Additionally, the results demonstrate that deeper network architectures consistently yield better segmentation performance for both anatomy and instruments, reflecting the increased capacity needed to model the complexity of these tasks. Notably, the SAM-LoRA model achieves the highest Dice scores across all segmentation classes.

Table~\ref{tab:quantitative-rhexis} summarizes the rhexis segmentation performance, addressing what is arguably the most challenging task in wet-lab skill assessment. The complexity stems from two factors: the limited number of annotated frames due to the short duration of the rhexis, and the inherently blunt edges of the rhexis boundary.
For this task, in addition to the network architectures evaluated for anatomical structures and instruments, we trained and evaluated the Segment Anything Model (SAM) using pupil bounding box prompts derived from the pupil segmentation results. As reported in the table, among networks operating without supervised prompts, SAM-LoRA achieved the highest performance. Notably, this performance is significantly enhanced when pupil bounding box prompts are incorporated, as reflected by the results for SAM-LoRA (bb).

\section{Conclusion}
\label{sec: conclusion}

In this paper, we introduce WetCat, the first curated dataset specifically designed for skill assessment in wet-lab cataract surgery videos. By providing comprehensive phase annotations and semantic segmentations focused on the critical capsulorhexis and phacoemulsification phases, WetCat enables the development of interpretable, AI-driven evaluation tools. This resource addresses a critical gap in ophthalmic surgical education, fostering advancements in scalable, objective, and consistent surgical training.

\begin{acks}
This work was funded by HaagStreit AG. We would like to thank Finia Kowal, Alessia Bruzzo, and Ylenia Di Maro for their invaluable contributions to the meticulous annotation of the dataset.

\end{acks}

\balance
\bibliographystyle{acm}
\bibliography{bibtex}

\end{document}